\documentclass[letterpaper, 10 pt, journal, twoside]{IEEEtran}
\IEEEoverridecommandlockouts

\usepackage{times}
\usepackage{epsfig}
\usepackage{graphicx}
\usepackage{amsmath}
\usepackage{amssymb}
\usepackage[noend]{algpseudocode}
\usepackage{balance}
\usepackage{color}
\usepackage{url}
\usepackage{authblk}
\usepackage{multirow}
\definecolor{mygray}{gray}{0.6}
\graphicspath{{figures/}}
\usepackage{tabularx}

\newcommand{\etal}{\emph{et~al.}}

\usepackage{cite}
\usepackage[linesnumbered,ruled]{algorithm2e}

\usepackage[flushleft]{threeparttable}

\makeatletter
\def\BState{\State\hskip-\ALG@thistlm}
\makeatother
\algdef{SE}[DOWHILE]{Do}{doWhile}{\algorithmicdo}[1]{\algorithmicwhile\ #1}

\def\cblue{\textcolor{black}}
\definecolor{black}{rgb}{0, 0, 0}
\newcommand{\sk}[2]{\textcolor{black}{{#1}} \textcolor{black}{{#2}}}
\newcommand{\skold}[1]{\textcolor{black}{{#1}}}
\newcommand{\sknew}[1]{\textcolor{black}{{#1}}}

\graphicspath{ {figures/} }

\def\creview{\textcolor{black}}
\def\camrdy{\textcolor{black}}

\newcolumntype{L}[1]{>{\raggedright\arraybackslash}p{#1}}
\newcolumntype{C}[1]{>{\centering\arraybackslash}p{#1}}
\newcolumntype{R}[1]{>{\raggedleft\arraybackslash}p{#1}}

\title{\LARGE \bf
Robust Photogeometric Localization over Time 
\protect\\ for Map-Centric Loop Closure\\
}

\author{Chanoh Park$^{1,2}$, Soohwan Kim$^{3}$, Peyman Moghadam$^{1,2}$, Jiadong Guo$^{1,4}$, Sridha Sridharan$^{2}$, Clinton Fookes$^{2}$ 
\thanks{Manuscript received: September, 10, 2018; Revised December, 14, 2018; Accepted January, 7, 2019.}
\thanks{This paper was recommended for publication by Editor Cyrill Stachniss upon evaluation of the Associate Editor and Reviewers' comments.}
\thanks{
$^1$ Chanoh Park, Peyman Moghadam are with the Robotics and Autonomous Systems Group, DATA61, CSIRO, Brisbane, QLD 4069, Australia.
E-mails: {\tt\footnotesize \emph{Chanoh.Park, Peyman.Moghadam}@data61.csiro.au} }
\thanks{
$^{2}$ Sridha Sridharan, Clinton Fookes are with the School of Electrical Engineering and Computer Science, Queensland University of Technology (QUT), Brisbane, Australia.
E-mails: {\tt\footnotesize \emph{chanoh.park, peyman.moghadam, s.sridharan, c.fookes}@qut.edu.au}}
\thanks{
$^{3}$ Soohwan Kim is with Division of Smart Automotive Engineering, Sun Moon University, South Korea, E-mail: {\tt\small \emph{kimsoohwan}@gmail.com}.}
\thanks{
$^{4}$
Jiadong Guo is with Autonomous Systems Lab, ETH Zurich, Switzerland. E-mail: {\tt\footnotesize \emph{jiadong.guo}@outlook.com}}
\thanks{Digital Object Identifier (DOI): see top of this page.}
}

\begin{document}

\maketitle

\begin{abstract}

Map-centric SLAM is emerging as an alternative of conventional graph-based SLAM for its accuracy and efficiency in long-term mapping problems. However, in map-centric SLAM, the process of loop closure differs from that of conventional SLAM and the result of incorrect loop closure is more destructive and is not reversible.
In this paper, we present a tightly coupled photogeometric metric localization for the loop closure problem in map-centric SLAM. In particular, our method combines complementary constraints from LiDAR and camera sensors, and validates loop closure candidates with sequential observations.
The proposed method provides a visual evidence-based outlier rejection where failures caused by either place recognition or localization outliers can be effectively removed. 
We demonstrate the proposed method is not only more accurate than the conventional global ICP methods but is also {robust to incorrect initial pose guesses.} 
\end{abstract}


\section{Introduction}
\label{sec:Introduction}

\IEEEPARstart{S}{imultaneous} Localization And Mapping (SLAM) is a key enabling component of autonomous robots and driverless technologies.
Although SLAM techniques have reached a mature level, with the recent advent of low cost multi-modal sensors, 3D dense LiDAR-based SLAM is still in its infancy.
\creview{
Map-centric approaches \cite{whelan2015, park2017c}, which have demonstrated their accuracy and effectiveness by fusion-based mapping and deformation-based loop closure, provide an
alternative solution to the dominant trajectory-centric SLAM \cite{bosse2012,vidas2015real}.}
\creview{
However, in
the map-centric approach, the effect of incorrect loop closure is more destructive and irreversible compared to the conventional graph-based methods. The map-centric approach fuses all the measurements on-the-fly to the map, where the fused map is difficult to be detached and corrected once the loop is closed. Therefore, the accuracy and robustness of loop closure in map-centric approaches \cite{whelan2015, park2017c} become crucial.}

\creview{
Previous map-centric approaches either did not have a proper method to detect false positives \cite{park2017c} in a loop closure or they are not robust enough as well as being limited to only the RGB-D sensor model \cite{whelan2015}.
The current state-of-the-art solution in false positive loop closure detection or localization failure detection \cite{ho2007} requires a global pose graph optimization for each
validity test which is a huge penalty when it comes to scalability issues. Furthermore,
their method is also dedicated for trajectory-centric systems such as graph-based SLAM, and as a result, is not applicable to map-centric methods.}

To overcome this problem, we propose a metric localization, namely \textit{PhotogeoSeq$^+$}, which combines geometric and photometric constraints over time for the robust metric localization of the map-centric loop closure.
The specific contributions of this work can be outlined as follows. First, with the tightly coupled \creview{geometric and photometric} constraints, our method robustly estimates 6 Degree of Freedom (DoF) \creview{alignment} even when the initial guess for registration is completely incorrect or unknown. 

Second, instead of making a loop closure decision based on only a single 6 DoF \creview{alignment} estimation, the proposed method observes the stability of the \creview{alignment} over time and provides a reliability metric to reject or accept a loop closure hypothesis without a global trajectory optimization.
\camrdy{For this purpose, we demonstrate that utilizing the visual features are beneficial
for detecting a localization failure.}

\camrdy{This paper is organized as follows. Section II reviews related works. Section III describes the overall procedure of the proposed method. In Section IV, we describe the proposed photogeometric metric localization method (\textit{PhotogeoSeq$^+$}), and detail our visual evidence-based outlier rejection in Section V.   Results demonstrating the advantages of our method over current state-of-the-art global ICP algorithms are given in Section VI. Finally we summarize and present future directions in Section VII. }

\section{Related Work}

SeqSLAM \cite{milford2012} proposed a place recognition architecture that utilizes the coherent sequential \creview{2D visual information, where it searches whether a similar sequence of the features between the two paths are observed over time.} 
Lynen \etal{}~\cite{lynen2014} extended SeqSLAM as a 2D probability density estimation problem in votes versus travel distance space. Their work finds chunks of revisited places more efficiently than SeqSLAM by the distance normalization.
However, as their method purely depends on 2D visual input, it is not suitable for a 6 DoF pose estimation problem. 
Furthermore, if the trajectory is only overlapped a short distance, such as at intersections, it is not likely to have much coherency which increases the risk of failure \cite{cummins2008}.

Latif \etal{}~\cite{latif2013} proposed a graph-based loop closure detection and correction system, where they utilize a residual of the graph optimization as a prior of a loop closure \camrdy{failure}. Their method clusters topologically related loop closure hypotheses over time and rejects any hypothesis that increases the residual of graph optimization. They have proved that the localization could be reversely utilized for testing whether the input place recognition is false positive, whereas the conventional algorithms showed single directional hierarchical flow from place recognition to localization. Although, they were able to increase robustness to false positive loop closures, their method requires a graph-optimization to test each hypothesis, which is a huge drawback for a \creview{long-term} mapping system.

On the other hand, the modality of the loop closure system is one of the key components \camrdy{that increases the performance and accuracy of a loop closure.} In early works, a single sensor was utilized for the loop closure detection problem, where 3D or visual descriptors were solely used \cite{cummins2008,latif2013}. 
However, with the single sensor approach the robustness is rather limited because of the innate degeneracies in each configuration. Thus, even the state-of-the-art works have a higher false positive rate than desired and work well only within limited scenarios.

There are few works that introduce multi-modality \creview{for} the place recognition or localization problem. Early works in this category \cite{ho2006,glocker2015,gawel2016}, combined 3D and 2D descriptors for a place recognition problem. 
While the combined complementary descriptors \creview{are capable of handling} degenerate configurations such as where there are no visual patterns or where the scene is geometrically flat, \creview{rejection of incorrect place recognition or alignment error is difficult.} Furthermore, the constraints from 2D recognition are not tightly integrated in the pose optimization of the metric localization step \cite{whelan2015,ho2006}. \camrdy{Also,} when it comes to a separate sensing system such as an independent LiDAR and camera system on a dynamically moving platform \cite{bosse2012}, 
\camrdy{the quality of the localization is affected by the spatio-temporal differences in observations,
which was not properly considered in the previous works \cite{ho2006}. }
Even in the cases where the integrated intensity from LiDAR is utilized instead of 2D visual features, difficulties still exist because of the LiDAR intensity calibration and intensity differences according to the incident angle \cite{guo2019}.

\section{Overview of the system}

\begin{figure}[t]
\centering{
\def\svgwidth{60mm}
\includegraphics[height=.14\textheight]{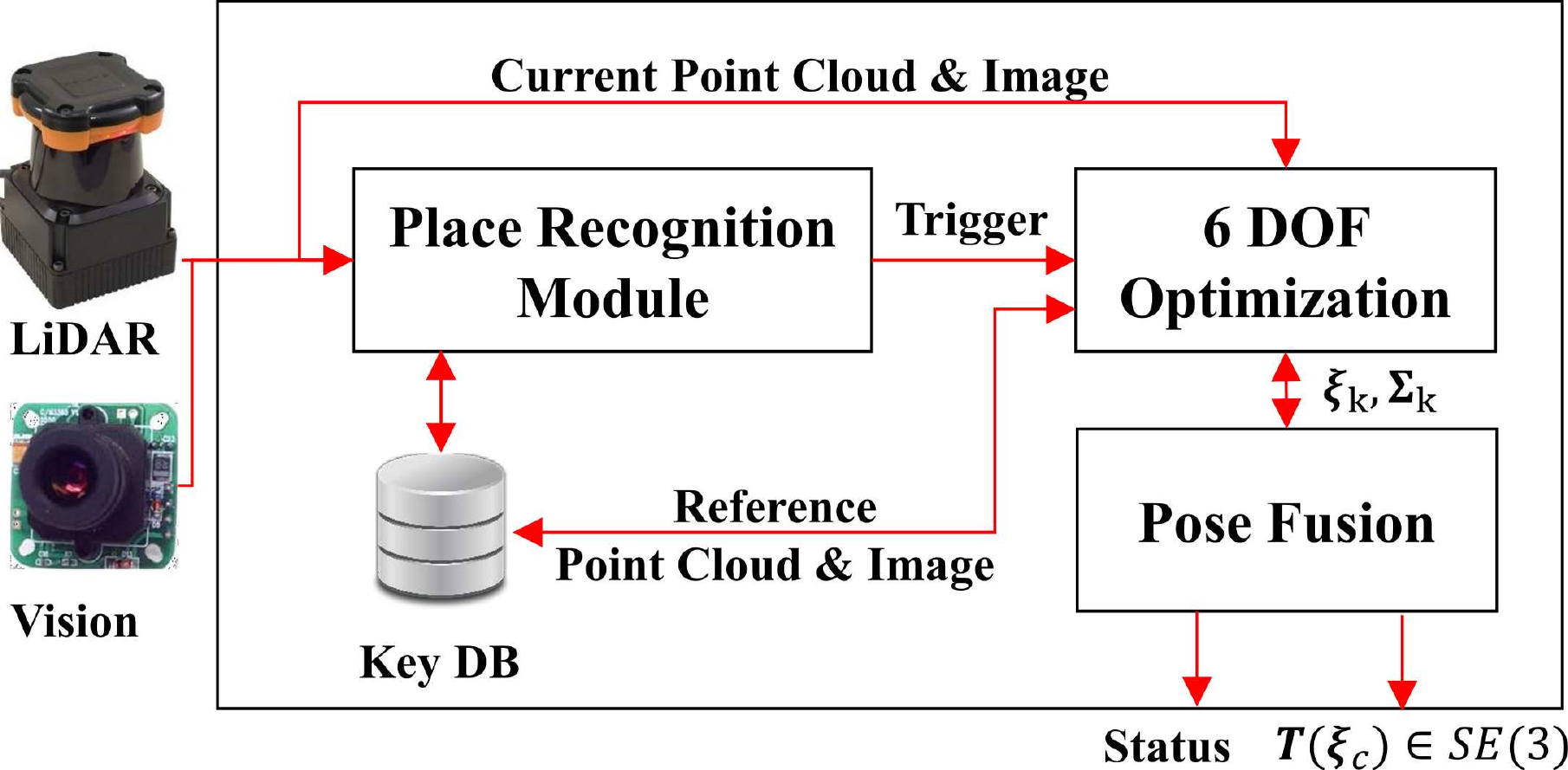}
}\vspace{-2mm}
\caption{
{\camrdy{Block diagram of the proposed \textit{PhotogeoSeq$^+$} method.}} 
}
\label{fig:blockdig}
\vspace{-2mm}
\end{figure}

\begin{figure}[t]
\begin{minipage}[t]{0.28\linewidth}
    \includegraphics[width=0.99\columnwidth]{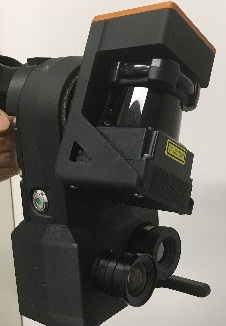}    
    \centering {\footnotesize {(a)} \normalsize }
\end{minipage}%
    \hfill%
\begin{minipage}[t]{0.7\linewidth}
    \includegraphics[width=1\columnwidth]{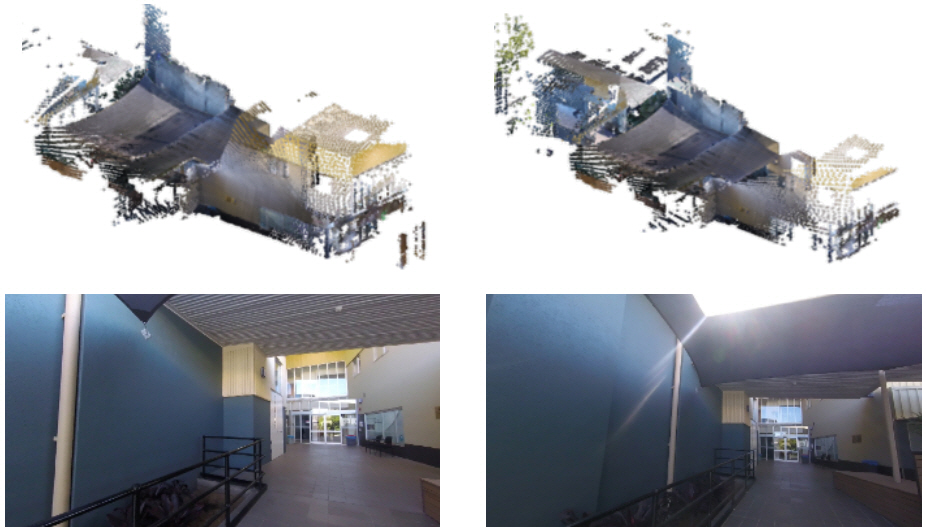}
    \centering {\footnotesize {(b)} \normalsize }
\end{minipage}
\caption{(a) A hand-held 3D scanning device with a spinning LiDAR, camera and IMU. (b) Pair of matched source and destination images from a place recognition module. The point clouds are extracted at the timestamp of each image.  }
\label{fig:device}  
\vspace{-4mm}
\end{figure}

\creview{The proposed metric localization method (\textit{PhotogeoSeq$^+$}) consists of {place recognition}, \creview{metric} localization, and pose fusion modules \camrdy{as depicted in Fig. \ref{fig:blockdig}.} 
Given a motion-undistorted local point cloud from LiDAR and corresponding camera image\creview{s} by the sliding window-based local trajectory optimization \cite{park2017c} as shown Fig. \ref{fig:device}, the {place recognition} module finds a possible revisit of a place and trigger{s} the localization step to estimate an alignment between the current location and the reference location. 
Once the localization module is triggered, our method continuously estimates the alignment between the new current location and its corresponding reference location at different places until the uncertainty of the estimated alignment \creview{reaches a certain} threshold.
}
\section{\creview{Alignment} Estimation}

In this section, we elaborate on our \creview{alignment}
estimation method by photogeometric constraints.

\setcounter{subsection}{0}
\vspace{-4mm}
\subsection{Continuous-Time Trajectory Representation}

\creview{In the proposed method, the synchronization problem of the multi-modal sensors such as different frame-rates or unsynchronized clocks are required to be properly addressed. To cope with this problem, our method
proposes to use the continuous-time trajectory representation \cite{park2017c}.} In the continuous-time trajectory representation, the trajectory is modeled as a function of time. An exact system pose $\textbf{T}(\tau) \in {SE(3)}$ at an arbitrary {query} time $\tau \in {\mathbb{R}}$ can be interpolated from a set of temporally nearby poses. Let $\textbf{T}(\tau)$ be composed of translational component $\textbf{t}{_\tau} \in {\mathbb{R}^3}$ and rotational component $\textbf{R}{_\tau} \in {SO(3) }$  as,
\begin{equation}
\textbf{T}(\tau):=\begin{bmatrix}
 \textbf{R}{_\tau}& \textbf{t}{_\tau} \\ 
\textbf{0} &1 
\end{bmatrix}.
\end{equation}

Then, utilizing the linear continuous-time trajectory representation, its value can be evaluated by an interpolation from two poses $ \textbf{T}{_k}, \textbf{T}{_{k+1}}  $ where their timestamp{s} satisfy $\tau_{k}<\tau<{\tau}_{{k+1}}$. Given poses and timestamps, their interpolation at $\tau$ is given by, 
\begin{equation}
\textbf{T}(\tau)= \textbf{T}{_{{k}}} \textbf{e}{^{\alpha[\boldsymbol{\xi}_r]{_\times}}}.
\end{equation}
where the relative pose $[\boldsymbol{\xi}_r]_\times\in\mathfrak{s}\mathfrak{e}(3)$ is defined by $log(\textbf{T}{_{{k}}^{-1}}\textbf{T}{_{{k+1}}})$ and  the exponential mapping $\textbf{e}{^{\alpha[\boldsymbol{\xi}]{_\times}}}$ linearly interpolates the relative pose on the manifold with the interpolation ratio $\alpha = ({\tau-{\tau}_{{k}}})/({\tau_{k+1}-{\tau}_{{k}}})$. \camrdy{$[\cdot]_{\times}$ represents skew-symmetric matrix conversion of a vector.}

\vspace{-4mm}
\subsection{{Alignment} Between Two Places}

\begin{figure}[t]
\centering{
\def\svgwidth{75mm}
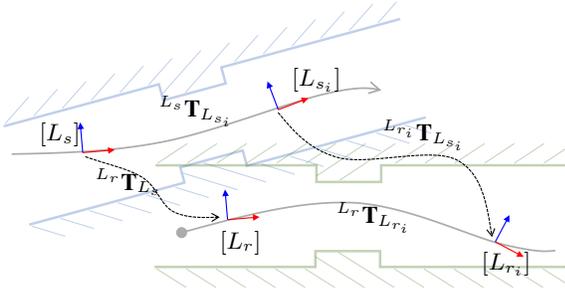
}
\caption{Illustration of the proposed 6 DoF \creview{alignment} estimation. Each \creview{alignment} $^{L_{r_i}}\textbf{T}_{L_{s_i}}$ between source [$L_{s_i}$] and reference [$L_{r_i}$] at different $i_{th}$ places are transformed to the very first place [$L_s$], [$L_r$] where the loop closure was first reported. In such way, what we keep estimating is $^{L_{r}}\textbf{T}_{L_{s}}$.  Selection of $i_{th}$ source [$L_{s_i}$] and reference [$L_{r_i}$] locations can be decided from multiple sources such as 2D or 3D place recognition and relative motion.
}
\label{fig:multipleLoc}       
\vspace{-4mm}
\end{figure}

\creview{Given a continuous-time trajectory and two matching places found by a loop closure detection module as depicted in Fig. \ref{fig:device} (b), we define an initial  source to reference transformation from the drifted raw trajectory as,}\looseness=-1 
\begin{equation}
\begin{split}
{}^{L_r}{\textbf{T}}_{L_s} = 
{}^{w} \textbf{T}_L(\tau_r)^{-1}  {}^{w} \textbf{T}_L(\tau_s),
\end{split}
\label{eq:src2dst} 
\end{equation}
\creview{where  $^{w} {\textbf{T}_{L}}(\tau)$ is a continuous-time trajectory \cite{bosse2012} that represents the LiDAR frame $[L]$ in the world frame $[w]$ at source and reference time $\tau_s$, $\tau_r$ as shown in  Fig. \ref{fig:multipleLoc}. 
Due to drifts in trajectory, the initial alignment ${^{L_r} \textbf{T}_{L_s}}$ is highly likely to be misaligned.
 Thus, we define and find its correction by,}
\begin{equation}
\begin{split}
{}^{L_r}\bar{\textbf{T}}_{L_s} = \boldsymbol{e}^{[\boldsymbol{\xi}]_\times} {}^{L_r}{\textbf{T}}_{L_s}
&:=\begin{bmatrix}
 \textbf{R}& \textbf{t} \\ 
\textbf{0} & 1 
\end{bmatrix},
\end{split}
\label{eq:disp} 
\end{equation}
\creview{where $[{\boldsymbol{\xi}}]_\times\in {\mathfrak{se}(3)}$ is the compensation to the initial alignment which can be found by visual and geometrical
sensor observations.}
\creview{Additionally, for the joint optimization by the LiDAR and visual observations,
we define a corresponding relative camera transformation as,}
\begin{equation}
\begin{split}
^{C_r} \textbf{T}_{C_s} &={^{C} \textbf{T}_{L}}{^{L_r} \textbf{T}_{L_s}} {^{L} \textbf{T}_{C}}
:=\begin{bmatrix}
 \textbf{R}'& \textbf{t}' \\ 
\textbf{0} & 1 
\end{bmatrix},
\end{split}
\end{equation}
\creview{where ${^{L} \textbf{T}_{C}}$ is a pre-calibrated camera frame [C] to LiDAR
frame extrinsic{s}.}

\subsection{Single Alignment Estimation from Place Pairs}

 As we already mentioned, an \creview{alignment} estimation based on a single location has a high probability of failure. We propose to aggregate \creview{the estimations
of multiple alignments} at different locations to the very first place where the loop closure was first detected. This approach will develop a more robust and reliable failure detection and outlier rejection. 
  
Equation (\ref{eq:disp}) represent\sknew{s} the displacement between two matched local frames. As our objective is to estimate a single alignment,
we transform the local displacements into the first place as, 
\begin{equation}
\label{eq:overtime}
{}^{L_r}{\textbf{T}}_{L_s}^i={^{\sknew{L}_r} \textbf{T}_{\sknew{L}_{r_i}}}{^{\sknew{L}_{r_i}} \textbf{T}_{\sknew{L}_{s_i}}}{^{\sknew{L}_{s_i}} \textbf{T}_{\sknew{L}_s}},
\end{equation}
where ${^{\sknew{L}_r} \textbf{T}_{\sknew{L}_{r_i}}}$, ${^{\sknew{L}_s} \textbf{T}_{\sknew{L}_{s_i}}}$ are the \creview{relative} transformation from the first frame and $i_{th}$ frame \creview{which
are extracted from the drifted trajectory} as described in Fig. \ref{fig:multipleLoc}. It is important to note that we assume that the trajectory utilized to extract ${^{\sknew{L}_r} \textbf{T}_{\sknew{L}_{r_i}}}$, ${^{\sknew{L}_{s_i}} \textbf{T}_{\sknew{L}_s}}$ is locally rigid and accurate enough, which\sk{}{implies} that
${^{\sknew{L}_r} \textbf{T}_{\sknew{L}_{r_i}}}$, ${^{\sknew{L}_{s_i}} \textbf{T}_{\sknew{L}_s}}$\sk{}{do not embed} any uncertainty because of the local trajectory drift within small regions.

\subsection{\sknew{Spatio-Temporal Uncertainty in Visual Measurements}}

\creview{To account for uncertainties} from various sources during the optimization, we define the uncertainty model of the system. \creview{Two types of uncertainties are defined}. The first uncertainty is from the temporal difference between the two \creview{sensing modalities}. As we estimate the location of the camera based on the LiDAR\ trajectory, the temporal uncertainty affects the camera location estimation as,  
\looseness=-1
\begin{equation}
^{w} \textbf{T}_{C} = {^{w} \textbf{T}_{L}}(\tau+\delta\tau){^{L} \textbf{T}_{C}} ,
\end{equation}
where $\delta\tau \sim\ \mathcal{N}(\mu_\sigma,\sigma_{\tau}^2)$ is the estimated temporal difference with mean $\mu_\sigma$ and variance $\sigma_{\tau}^2$ \creview{\cite{park2018}}. Even with a small temporal uncertainty, a large local motion amplifies the uncertainty of the camera pose. 
\begin{figure}[t]
\centering{
\def\svgwidth{55mm}
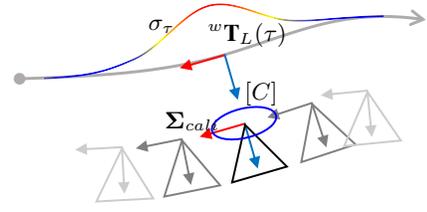
}
\caption{
LiDAR trajectory and LiDAR pose ${}^{w} \textbf{T}_L(\tau)$ at camera image time stamp $\tau$. Actual camera position can be any where around the given time stamp. The Gaussian distribution on the trajectory represents the time lag variance region. Spatial uncertainty caused by calibration error is represented as the blue circle.
}
\label{fig:timelag}       
\vspace{-4mm}
\end{figure}

The second uncertainty originates from the extrinsic calibration between the \sknew{two} sensors,\skold{}
\begin{equation}
{^{L} \textbf{T}_{C}}=\boldsymbol{e}^{[\boldsymbol{\xi}_{cali}]_\times}{^{L} \bar{\textbf{T}}_{C}},
\end{equation}
where ${\boldsymbol{\xi}_{cali}} \sim\ \mathcal{N}(\textbf{0},\boldsymbol{\Sigma}_{cali})$ is \creview{an extrinsic} calibration error and  ${^{L} \bar{\textbf{T}}_{C}}$ is the true extrinsic\sknew{s}. The visualization of these uncertainties are shown in Fig. \ref{fig:timelag}.
\sknew{The uncertainties modelled here will be propagated through the cost function of visual constraints and used as weights in non-linear least squares in the next section} as well as a statistic test for localization outlier rejection. 

\vspace{-4mm}
\subsection{Geometric Constraints} 
\sk{}{We apply surfel-based point-to-plane ICP with multi-resolutional voxelization\cite{park2017b}} for the geometrical alignment between the source and the reference point cloud as,
\begin{equation}
\label{eq:ICP}
\textbf{e}{_{I}}=\textbf{n}^\top(\textbf{p}_r-(\textbf{R}\textbf{p}_s+\textbf{t})),
\end{equation}
\begin{equation}
\label{eq:surfelICP}
\mathbf{r}_{I}=\frac{1}{2}\sum_{v=1}^{V}{\textbf{e}{_{I_v}}^\top\boldsymbol{\Sigma}_{I_v}^{-1}\textbf{e}{_{I_v}}},
\end{equation}
where the pair of the matched surfel centroids $\textbf{p}_r$, $\textbf{p}_s$\sk{}{are} acquired by a nearest neighbour method in \sknew{the} surface normal and centroid space. The original point cloud to calculate these surfels are extracted around the pose $^{w} \textbf{T}_{L}(\tau)$ at $\tau_s$, $\tau_r$ and represented in the local LiDAR coordinate\sknew{s}. 
The covariance is defined by a M-estimator weight and the Eigenvalue of the voxel \cite{park2017c}.

\subsection{Photometric Constraints} 
\label{eq:ddddd}

\creview{The sparse surfel constraint is computationally efficient because of the lower number of surfels after a spatial compression but this often fails when the initial guess on the pose is incorrect or the scene is geometrically degenerate such as in a corridor or wide open area.} To cope with both problems, we tightly combine Photometric constraints with ICP. If {there are distinctive visual patterns in the scene}, the visual patterns generate complementary constraints which are not effected by {the structure} of the scene.

Our proposed method combines two different types of visual feature{s}: semi-{direct} features \cite{forster2014} and {indirect} features \cite{lowe2004}, as a complementary constraint to the geometric constraint. Both visual features are fed into {the} epipolar constraint\skold{s} between two camera poses. Thus, given a set of 2D features $\textbf{u}_s$, $\textbf{u}_r$  the epipolar constraint is given as,  
\begin{equation}
\label{eq:epi}
\textbf{e}{_{c}}={\textbf{u}_s^\top} [\textbf{t}']_{\times}\textbf{R}'\textbf{u}_r .
\end{equation}
The visual constraint by (\ref{eq:epi}) can estimate the pose up to scale only. However, combined with \sknew{the} ICP constraint, full 6 DoF can be estimated as well as preventing the optimization from\sk{}{being stuck} at\skold{} local minima because of\skold{} incorrect surfel matching caused by \skold{}an incorrect initial guess on alignment.\looseness=-1

The \sknew{in}direct 2D features between the source image and the reference image are found by a global feature descriptor method \cite{lowe2004} to cope with \sknew{the} large view difference problem. 
As the feature detection and matching stage are independent of \sknew{the} time lag and initial guess on the \creview{true alignment}, even\sk{}{a few matches impose} strong orientational constraints. This will guide the pose estimation to the roughly correct direction so that ICP can find \sknew{the} proper {matchings of surfels}{}. Therefore, Equation (\ref{eq:epi}) largely reduces the chance of getting stuck in local minima {because of surfel false matchings}{}.      

Also, to estimate more features widely spread over the image, the semi-direct features are matched by projecting \sknew{the} reference image feature\sk{}{points} onto the source and refining as depicted in Fig. 5.
It starts by defining the\sk{}{warping} function\sk{}{$\mathcal{W}(\textbf{u}_r)$} as,
\begin{equation}
\label{eq:direct}
\textbf{u}_s'=\mathcal{W}(\textbf{u}_r)=\boldsymbol{\pi}(\textbf{K}(\textbf{R}'\textbf{p}(\textbf{u}_r,\mathcal{D}_r)+\textbf{ t}'),
\end{equation}
where\sk{}{$\textbf{u}_r$, $\textbf{u}_s' \in R^2$} are\sk{}{the feature location on the reference image and its projection on the source image}, {$\textbf{K}$ is the intrinsic matrix} and $\mathcal{D}_r$ is the depth map of the scene, and $\pi$ is the camera projection. {$\textbf{p}(\textbf{u}_r,\mathcal{D}_r)\in \mathbb{R}^3$ is the reconstructed 3D position{s} of the 2D feature{s}.} The depth map \sk{$\mathcal{D}_r$}{} is generated from a dense surfel map by the LiDAR point cloud \cite{park2017c} by rendering at ${^{w} \textbf{T}_{L}}(\tau){^{L} \textbf{T}_{C}} $ where $\tau$ follows the reference image time stamp.

\begin{figure}[t]
\centering{
\def\svgwidth{85mm}
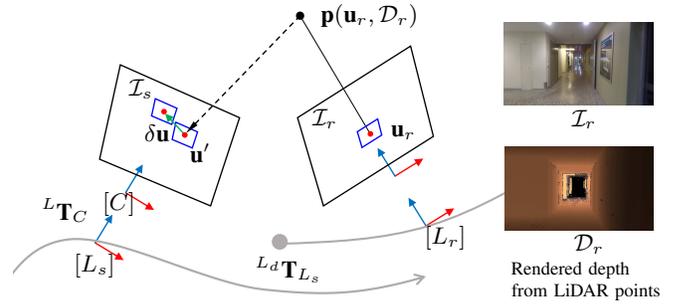
}
\caption{Illustration of the projected feature. The original feature location in the reference frame \cblue{${^{w} \textbf{T}_{C_r}}$} is reconstructed and projected onto the source frame \cblue{${^{w} \textbf{T}_{C_s}}$}. Then, the correction is found toward the direction that reduces the intensity difference of the two image patches. \camrdy{The depth map  $\mathcal{D}_r$ of the reference image $\mathcal{I}_r$ is achieved by rendering the surfel map.}   
}
\label{fig:projected_patch}       
\vspace{-2mm}
\end{figure}

\begin{figure}[t]
\centering{
\def\svgwidth{88mm}
\begingroup%
  \makeatletter%
  \providecommand\color[2][]{%
    \errmessage{(Inkscape) Color is used for the text in Inkscape, but the package 'color.sty' is not loaded}%
    \renewcommand\color[2][]{}%
  }%
  \providecommand\transparent[1]{%
    \errmessage{(Inkscape) Transparency is used (non-zero) for the text in Inkscape, but the package 'transparent.sty' is not loaded}%
    \renewcommand\transparent[1]{}%
  }%
  \providecommand\rotatebox[2]{#2}%
  \newcommand*\fsize{\dimexpr\f@size pt\relax}%
  \newcommand*\lineheight[1]{\fontsize{\fsize}{#1\fsize}\selectfont}%
  \ifx\svgwidth\undefined%
    \setlength{\unitlength}{719.66798797bp}%
    \ifx\svgscale\undefined%
      \relax%
    \else%
      \setlength{\unitlength}{\unitlength * \real{\svgscale}}%
    \fi%
  \else%
    \setlength{\unitlength}{\svgwidth}%
  \fi%
  \global\let\svgwidth\undefined%
  \global\let\svgscale\undefined%
  \makeatother%
  \begin{picture}(1,0.2286593)%
    \lineheight{1}%
    \setlength\tabcolsep{0pt}%
    \put(0,0){\includegraphics[width=\unitlength]{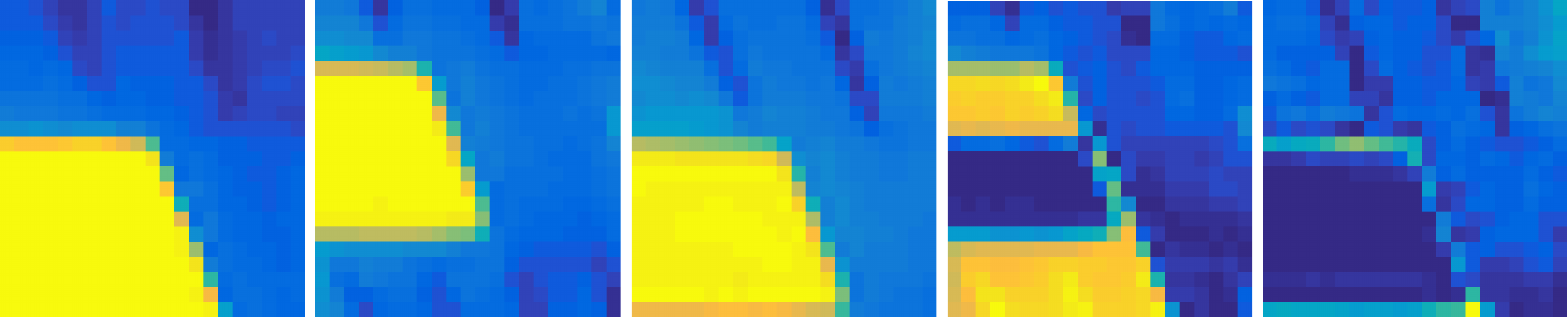}}%
    \put(0.06131129,-0.03085062){\color[rgb]{0,0,0}\makebox(0,0)[lt]{\lineheight{1.25}\smash{\begin{tabular}[t]{l}\fontsize{9pt}{1em} \footnotesize(a)\end{tabular}}}}%
    \put(0.26563584,-0.02979788){\color[rgb]{0,0,0}\makebox(0,0)[lt]{\lineheight{1.25}\smash{\begin{tabular}[t]{l}\fontsize{9pt}{1em} \footnotesize(b)\end{tabular}}}}%
    \put(0.46226165,-0.02979794){\color[rgb]{0,0,0}\makebox(0,0)[lt]{\lineheight{1.25}\smash{\begin{tabular}[t]{l}\fontsize{9pt}{1em} \footnotesize(c)\\\end{tabular}}}}%
    \put(0.86851768,-0.03023398){\color[rgb]{0,0,0}\makebox(0,0)[lt]{\lineheight{1.25}\smash{\begin{tabular}[t]{l}\fontsize{9pt}{1em} \footnotesize(e)\end{tabular}}}}%
    \put(0.66577732,-0.03023398){\color[rgb]{0,0,0}\makebox(0,0)[lt]{\lineheight{1.25}\smash{\begin{tabular}[t]{l}\fontsize{9pt}{1em} \footnotesize(d) \end{tabular}}}}%
  \end{picture}%
\endgroup%

}
\vspace{-4mm}
\caption{
\camrdy{Illustration of a matched patch refinement step. (a) Warped reference patch, (b) Source patch, (c) Shifted patch, (d) Initial error, (e) Error after refinement. Because of the time lag, calibration error and pose estimation error, a projected and warped reference patch does not exactly matches with the source patch location. The refinement step corrects the difference.} }
\label{fig:patchwarp}       
\vspace{-4mm}
\end{figure}
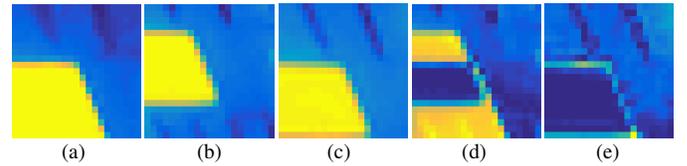

As the depth is loosely coupled to the image, the 2D geometry of the projected reference patch is not accurately aligned to the source patch in general
as shown in the Fig. \ref{fig:patchwarp}. Thus, for each image patch a refinement $\delta\textbf{u}^*$ is found so that it reduces the intensity difference of two patches as described in Fig. 5 and Fig. \ref{fig:patchwarp},
\begin{equation}
\delta\textbf{u}^*=\underset{\delta\textbf{u}}{\mathrm{argmin}}{\sum_{\Delta\textbf{u}\in \mathcal{P}}{ \Bigl\| \mathcal{I}_s(\textbf{u}_s'+\delta\textbf{u}+\Delta\textbf{u})-\mathcal{I}(\Delta\textbf{u})} \Bigl\|^2},
\label{eq:ctobj}
\end{equation}
where $\mathcal{I}_s$, $\mathcal{I}$ are the source and reference images defined at $\textbf{u}_s'$ respectively, $ \mathcal{P}$ is the image patch around $\textbf{u}_s'$. The warped reference image $\mathcal{I}$ is defined as, 
\begin{equation}
\mathcal{I}(\Delta\textbf{u})=\mathcal{I}_r(\mathcal{W}^{-1}(\textbf{u}_s'+\Delta\textbf{u})),
\end{equation}
where $\mathcal{W}^{-1}(\textbf{u}_s')$ is the inverse projection from the source to the reference. Thus, the additional matched points are given from the reference feature location and the corrected source location as $\textbf{u}_r,\textbf{u}_s=\textbf{u}_s'+\delta\textbf{u}$.
Note that the semi-direct method does not directly relate the intensity to the pose estimation problem. The semi-direct method is utilized to find more matched points that are uniformly distributed over the image. This semi-direct method is inspired by SVO \cite{forster2014} but we added frequency domain refinement to cope with considerably large $\delta\textbf{u}$ \cite{guizar-sicairos2008}.

The\sk{}{semi-direct and indirect features} are also complementary to each other. 
The semi-direct method finds features only from the reference side and warps them to the source side. Although more features are found evenly over the entire image area, it is not functional until the given pose is close enough to the true pose. However, the matching of indirect features are given regardless of the current pose, therefore, they are valid over all pose estimation stages. Furthermore, the indirect features are usually detected at a far distance where the shape of features are less distorted, thus it is easy to become a rank deficient constraint.
Finally, the cost function of the two visual constraints are defined as,
\begin{equation}
\label{eq:surfelmat_map}
\mathbf{r}_{P}=\frac{1}{2}\sum_{j=1}^{J}{\textbf{e}_{c_j}^\top\boldsymbol{\Sigma}_{c_j}^{-1}\textbf{e}{_{c_j}}},
\end{equation}
where \camrdy{${e}_{c_j}$ is the error of $j_{th}$  feature pair} and the covariances $\boldsymbol{\Sigma}_{c_j} \in \mathbb{R}$ are utilized to consider spatio-temporal uncertainty caused by heterogeneous sensory system which is defined as,  
\begin{equation}
\begin{split}
\boldsymbol{\Sigma}= &{\mathbf{J}_{s}^\top}{\sigma_{p}}{\mathbf{J}_{s}}+{\mathbf{J}_{r}^\top}{\sigma_{p}}{\mathbf{J}_{r}}+{\mathbf{J}_{\tau_1}^\top}{\sigma_{\tau}}{\mathbf{J}_{\tau1}}\\
&+{\mathbf{J}_{\tau_2}^\top}{\sigma_{\tau}}{\mathbf{J}_{\tau_2}}+2 \mathbf{J}_{\boldsymbol{\xi}_{cali}}^\top\boldsymbol{\Sigma}_{cali} \mathbf{J}_{\boldsymbol{\xi}_{cali}},
\end{split}
\label{eq:uncertainty}
\end{equation}
where $\mathbf{J}_{s}$, $\mathbf{J}_{r}$, $\mathbf{J}_{\tau_{1,2}}$, $\mathbf{J}_{\boldsymbol{\xi}_{cali}}$ are the Jacobian of the cost function with respect to the uncertainty parameters on matching corruption, $\tau$ and $\boldsymbol{\xi}_{cali}$. They propagate the uncertainty to the optimization parameter space and are defined as follows,
\begin{equation}
\mathbf{J}_r=\frac{\partial \textbf{e}}{\partial \textbf{u}_r} ,\mathbf{J}_s=\frac{\partial \textbf{e}}{\partial \textbf{u}_s} ,\mathbf{J}_\tau=\frac{\partial \textbf{e}}{\partial \tau} , \mathbf{J}_{\boldsymbol{\xi}_{cali}}=\frac{\partial\textbf{e}}{\partial \boldsymbol{\xi}_{cali}}.
\end{equation}

The amount of \creview{the pixel level uncertainty $\sigma_{p}$ is calculated from a predefined ground truth dataset by back projecting source inlier features to the destination. The spatiotemporal uncertainties $\sigma_{\tau}$, $\boldsymbol{\Sigma}_{cali}$ are approximated
by the Hessian during the on-the-fly extrinsic, temporal calibration \cite{park2018} by using a continuous-time bundle adjustment. }

\vspace{-4mm}
\subsection{Joint Optimization} 
Given a photogeometric cost function $\mathbf{F}(\boldsymbol{\xi})= \alpha\mathbf{r}_{I}+\beta\mathbf{r}_{P}$, we {utilize} the Gauss-Newton {method} to find the \creview{compensation $\boldsymbol{\xi}$  and update ${^{\sknew{L}_{r_i}} \textbf{T}_{\sknew{L}_{s_i}}}$ iteratively}. 
 \creview{Since the cost function is a multi objective problem, a proper handling of scale difference is required \cite{munoz2016}. The scaling factors $\alpha, \beta$
are found by normalization.} {Also, for soft outlier rejection, we applied an M-estimator with t-distribution based weights.}

\section{Fusing Alignment Estimations}
 
Generally, single \creview{alignment} estimation is prone to error due to various inevitable reasons. In our proposed system, to ensure robust \creview{localization}, we propose to use sequential \creview{alignment} estimation through different parts of the map instead of taking a single place recognition result for a loop closure. 
In this section we describe our probabilistic and temporal approach on fusing estimated poses.
Also, an effective way to detect a localization failure or false positive will be discussed as well.\looseness=-1

\vspace{-4mm}
\subsection{Pose Fusion Model}

Suppose that we have multiple \creview{alignment} estimations and their covariances. As the estimated poses \creview{${}^{L_r}{\textbf{T}}_{L_s}^i$} are the  \creview{alignments} between the first source and reference frame, the \creview{estimated} alignment\sk{}{should} not change over time under the assumption that they are locally rigid. Thus, we model this problem as estimating the same pose over time by (\ref{eq:overtime}).     

The Bayesian fusion offers a closed form solution on the vector fusion problem \cite{park2017c}. However, as the pose vector sits on a manifold, directly applying the Bayesian fusion on the poses causes the convergence to be sub-optimal. Thus, we introduce the \skold{$SE(3)$} pose fusion approach proposed in \cite{barfoot2014} with a modification to convert the original batch fusion problem to a sequential fusion problem \creview{suitable for real-time application}.

Let $\textbf{T}_k,\boldsymbol{\Sigma}_k,\textbf{T}_c,\boldsymbol{\Sigma}_c$, respectively be $k_{th}$ pose measurement and the current pose estimation up to the point \sknew{with their uncertainties}. The uncertainty of the current estimation is approximated by the Hessian as $\boldsymbol{\Sigma}_k=\sigma^2(\textbf{J}^\top \textbf{J})^{-1}$.  Applying \sknew{the} Gauss-Newton method with the two poses, we have, 
\begin{equation}
\begin{split}
\textbf{A}& =\boldsymbol{\mathfrak{I}}^{-T}_c\boldsymbol{\Sigma}^{-1}_c\boldsymbol{\mathfrak{I}}^{-1}_c+\boldsymbol{\mathfrak{I}}^{-T}_k\boldsymbol{\Sigma}^{-1}_k\boldsymbol{\mathfrak{I}}^{-1}_k ,\\
\textbf{b}& =\boldsymbol{\mathfrak{I}}^{-T}_c\boldsymbol{\Sigma}^{-1}_c\boldsymbol{\xi}_c+\boldsymbol{\mathfrak{I}}^{-T}_k\boldsymbol{\Sigma}^{-1}_k\boldsymbol{\xi}_k ,\\
\boldsymbol{\xi}& =\textbf{A}^{-1}\textbf{b},
\end{split}
\end{equation}
where \creview{$\boldsymbol{\mathfrak{I}}^{-1}$ is the Baker-Cambell-Hausdorff approximation \cite{barfoot2014} which is calculated from $\textbf{T}_k,\textbf{T}_c$. Refer to \cite{barfoot2014}
for a detailed derivation of the above.} The next current pose estimation is found by iteratively updating the best guess from the current poses as, 
\begin{equation}
\textbf{T}^* \leftarrow  \textbf{e}^{[\boldsymbol{\xi}]_{\times}}\textbf{T}^* ,
\end{equation}
where {the pose $\textbf{T}^*$ become\sknew{s} the new $\textbf{T}_c$ at the end of the iteration for the next fusion.} The covariance of the next current pose estimation is decided as follows at the end of the iteration,
\begin{equation}
\boldsymbol{\Sigma}_c = (\boldsymbol{\mathfrak{I}}^{-T}_c\boldsymbol{\Sigma}_c^{-1}\boldsymbol{\mathfrak{I}}^{-1}_c+\boldsymbol{\mathfrak{I}}^{-T}_k\boldsymbol{\Sigma}_k^{-1}\boldsymbol{\mathfrak{I}}^{-1}_k)^{-1}.
\end{equation}

\creview{Note that the accuracy of the sequential fusion is identical with
the batch fusion method but sequential fusion reduces unnecessary redundant
fusions.}
To decide whether or not to collect more evidence for the loop closure, we inspect the eigenvalues of the current pose estimation covariance and stop estimating the \creview{alignment} when \creview{eigenvalues} are sufficiently small such that $\sum\lambda( \boldsymbol{\Sigma}_c)_i< \theta_{th}$ for $i = [1,...,6]$, where $\theta_{th}$ is a sufficiently conservative threshold and $\lambda( \boldsymbol{\Sigma}_c)_i$ $i_{th}$ eigenvalue of $\boldsymbol{\Sigma}_c$.

\vspace{-4mm}
\subsection{Outlier Rejection}
The basic assumption for the pose fusion is that the uncertainty of the estimated pose is consistent and well represents the true error. A pose with \sknew{a} large error is fine to be fused as long as its covariance value is high. In such a case, the incorrect pose merely changes the fused pose when the system has the strong evidence that \sknew{the} previous pose is correct.  But in general it is not true because of blindness of ICP oriented covariance with respect to translation and possibility of degeneracy in 2D scenes. Thus, pose outliers \creview{where covariance does not properly represent its estimation quality} should be filtered prior to the fusion. 

Under the local rigidity assumption, any new\sknew{ly} estimated alignment should not be increasing the residual of previously constructed constraints. 
The outlier rejection stage utilizes this rigidity assumption to detect and properly handle any outlier\sknew{s} before\sk{}{fusing} them to our final alignment \sknew{estimation}.

For this stage, only \sknew{the} epipolar constraint\skold{}\sk{}{is} utilized as\skold{} its residual\sknew{s} proportionally increase\skold{s} when \sknew{the} input pose is incorrect\sk{.}{Note that the \camrdy{surfel-based} geometric feature matching does not follow this property.} The \camrdy{matched surfels are} found in normal and centroid spaces, therefore, what it implies is an abstracted spatial alignment rather than feature level correspondence. 
Thus, given a new pose to be tested, \sknew{a} statistic test using \sknew{the} epipolar constraint is carried out as follows,
\begin{equation}
\label{eq:outlier_rejection}
\sum_{l=1}^{L}{}\sum_{j=1}^{J_i}{}{\textbf{e}{_{jl}}(\boldsymbol{\xi}_{jl})^\top\boldsymbol{\Sigma}_{jl}(\boldsymbol{\xi}_{jl})^{-1}\textbf{e}{_{jl}}(\boldsymbol{\xi}_{jl})}<\chi_{\delta,0.95}^2 ,
\end{equation}
where $\textbf{e}{_{jl}}$ is residual of $j_{th}$ epipolar constraint of $l_{th}$ pose estimation with respect to the pose $\boldsymbol{\xi}_{jl}$ and $\delta$ is degree of freedom. The pose $\boldsymbol{\xi}_{jl}$ should be reversely calculated from the given pose estimation using the relationship between the first pose and $l_{th}$ pose as in (\ref{eq:overtime}). This test checks if the residuals from each previous pose estimations are within 95\%\ $\chi^2$ confidence area considering uncertainties caused by \sknew{the} time lag and \creview{extrinsic} calibration error.

\section{Experiment}
\label{sec:experiments}
In this section, we demonstrate the accuracy and robustness of the proposed method by comparing it to ground truth and other 3D point cloud based global localization methods.

\vspace{-4mm}
\subsection{Implementation Details}

\begin{table*}[t] 
\caption{Localization Accuracy Comparison.}
\vspace{-4mm}
{
\begin{center}
\begin{tabular}{l C{1.88cm}C{1.88cm}C{1.88cm}C{1.88cm}C{1.88cm}C{1.88cm}}
\hline\noalign{\smallskip}
Initial  & {(a)}  & {(b) } &{(c) }&{(d)}& {(e)}  & {(f)}  \\
 \cline{2-7}
 \noalign{\smallskip}
Guess&  Sparse ICP&Visual+ICP&SHOT+ICP&  FPFH+ICP& \textit{PhotogeoSeq}& \textit{PhotogeoSeq$^+$}\\
\noalign{\smallskip}\hline\noalign{\smallskip}
  Easy&  \textbf{0.04}(0.01)&  0.05(0.01)&      0.15(0.008)&   0.30(0.03)& 0.09(0.004)&  0.05(\textbf{0.003})\tabularnewline
  Medium&  0.40(0.19)&  0.12(0.11)&     0.13(0.005)&  1.64(0.30)& 0.09(\textbf{0.004})&  \textbf{0.04}(\textbf{0.004})\tabularnewline
  Hard&  2.52(2.42)&  1.49(0.71)&    0.17(0.005)&   13.8(0.10)& 0.10(\textbf{0.004})&  \textbf{0.04}(\textbf{0.004})\tabularnewline 
  
  \noalign{\smallskip}\hline\noalign{\smallskip}
  Time (sec)& 0.17&0.63&1.05&3.25&0.75$\times i_{n}$&2.04$\times i_{n}$\tabularnewline
\noalign{\smallskip}\hline  
\end{tabular}
\end{center}
}
\vspace{-2mm}
\begin{tablenotes}
  \small
  \item  {\footnotesize The estimations are compared to the ground truth to calculate error norm of translation ${e_\textbf t}$ and rotation (${e_\textbf r}$). Units are meter and rotation vector norm. For each noise level, initial poses are randomly generated according to the following parameters:   Easy $\sigma_{\theta=0.1},\sigma_{t=0.5}$, Medium $\sigma_{\theta=1}, \sigma_{t=5}$, Hard $\sigma_{\theta=10},\sigma_{t=50}$. $i_{n}$ is the number of the matched places in the sequence. \camrdy{The proposed method in (e) \textit{PhotogeoSeq} combines photogeometric constraints in sequential manner with the additional visual features from the semi-direct method in (f) \textit{PhotogeoSeq$^+$}. \normalsize }} 

\end{tablenotes}
\label{tbl:noise}
\vspace{-1mm}
\end{table*}

\begin{figure*}[t]
\centering{
\def\svgwidth{180mm}
\input{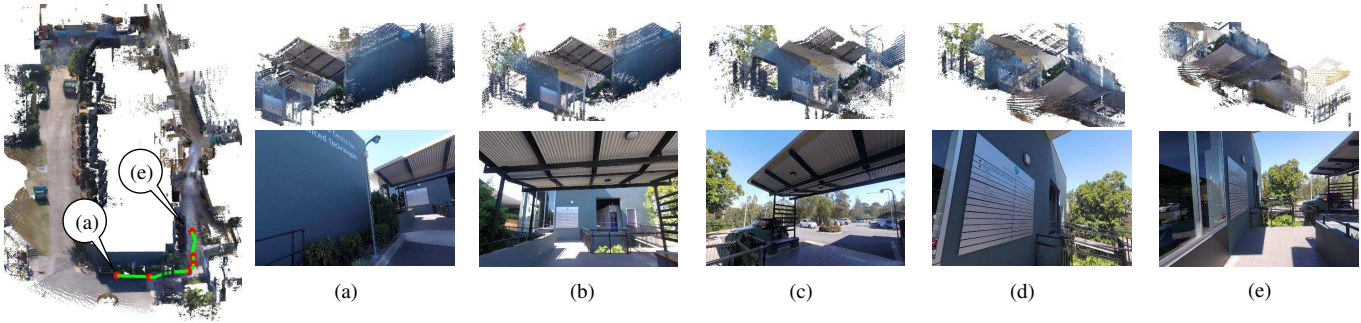}}
\vspace{-3mm}
\caption{
\camrdy{An example of a photogeometric localization sequence. The loop closure was first reported at (a) and finalized at (e).}
}
\label{fig:st_map}       
\vspace{-1mm}
\end{figure*}

For the experiments, a hand-held 3D spinning LiDAR is utilized to \creview{collect LiDAR and color images as shown in Fig. \ref{fig:device} (a).} \creview{The device consists of a spinning Hokuyo UTM-30LX laser, an encoder, a Microstrain 3DM-GX3 IMU, and a Grasshopper3 2.8 MP color camera\cite{park2017c}. }

For an initial seeding of the loop closure, a bag-of-word based place voting method is utilized. Once an initial pair of images are given, point clouds around the images are extracted for the ICP as shown Fig. \ref{fig:device} (b). \creview{To prevent the point cloud in the reference place and the current place to be mixed}, a temporal selection is utilized with $\pm$5 seconds range. \cblue{Also, for the efficiency in processing the point cloud, a 10$m$ spatial selection is made.} Additional matched place pairs are added by either the place recognition module or selecting the nearest image frame with a similar angle when an inlier pose exists. For the multi-resolutional voxels, voxels size with 0.3$m$, 0.8$m$, 1.5$m$ are utilized.

To reduce computational complexity, sub-sampled images with size 960$\times$540 are utilized with 21$\times$21 patch sizes for the semi-direct features.
Considering intensity changes caused by largely different views and time differences, the intensities of the patches are compared after a normalization.
For both the global and projected features, outliers are initially filtered using a geometrical constraint with the Essential matrix. Although, Essential matrix based filtering works reasonably well, it fails to detect outliers \creview{when 2D features are degenerate such as in co-planer cases.} 
Thus, upon \creview{detection of} an inlier 6 DoF localization that minimizes both the geometrical and visual difference within the expected uncertainty, all of the past 2D \creview{matches}, that are used as evidence in (\ref{eq:outlier_rejection}), are double checked and filtered \creview{by a $\chi^2$ test}. 
The localization estimation is terminated early when the uncertainty is sufficiently small.
\vspace{-2mm}
\subsection{\camrdy{Experimental Setup}}

\creview{As we mentioned earlier the initial guess on the true alignment is the dominant factor that often leads the pose optimization to failure.}
It is fair to assume the initial guess is unknown because of possible drift to a large extent. This is rather similar to the global point cloud registration problem. Thus, we compare our method with the state-of-the-art global ICP registration libraries with different initial guesses.  

\begin{figure*}[htbp]
\begin{minipage}[t]{0.28\linewidth}
    \includegraphics[width=0.99\columnwidth]{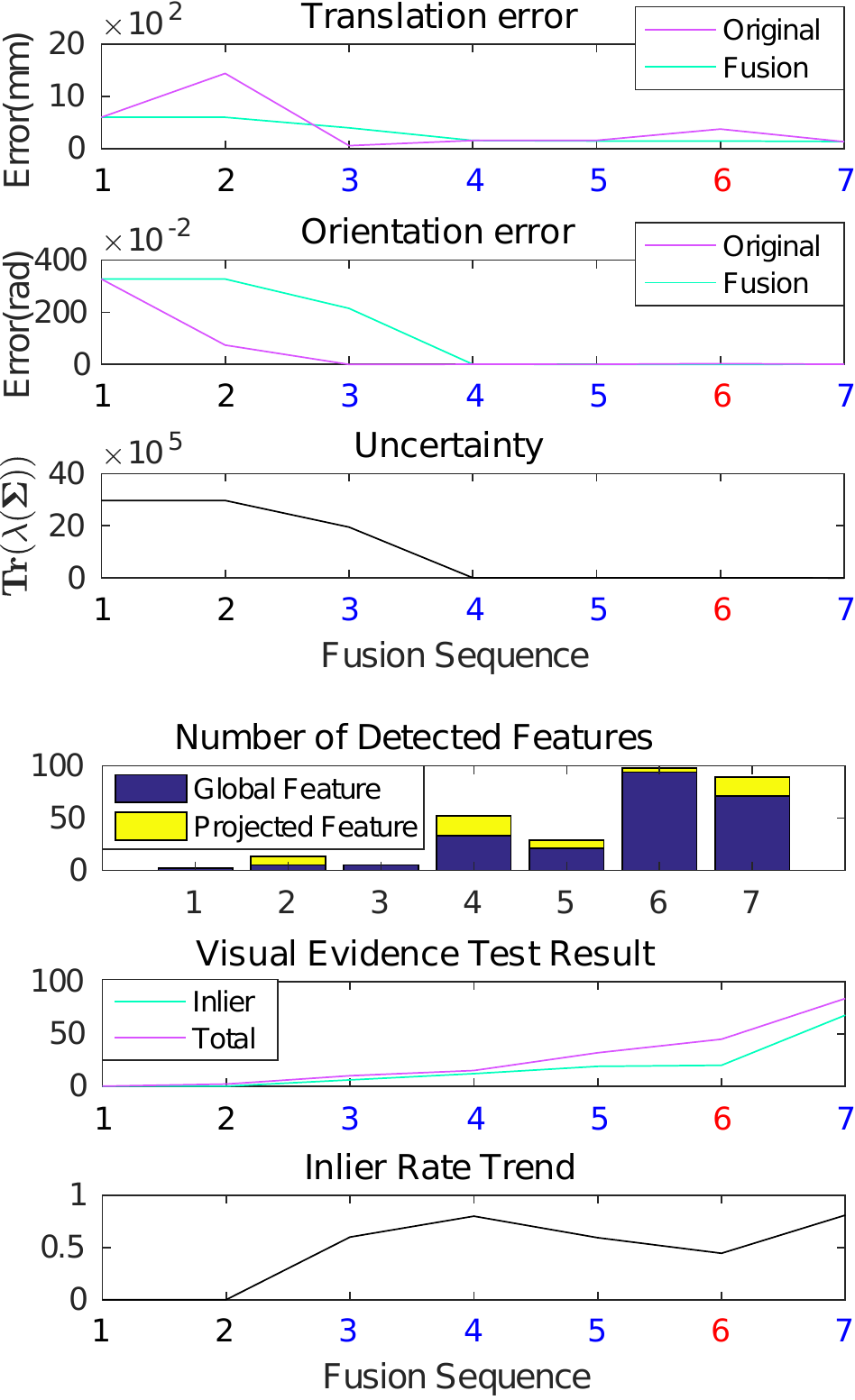}
    \caption{\creview{An example of pose error, uncertainty trends (upper) and visual evidence
test over fusions (lower).}}
    \label{fig:multi_fig}
\end{minipage}%
    \hfill%
\begin{minipage}[t]{0.7\linewidth}
    \includegraphics[width=1\columnwidth]{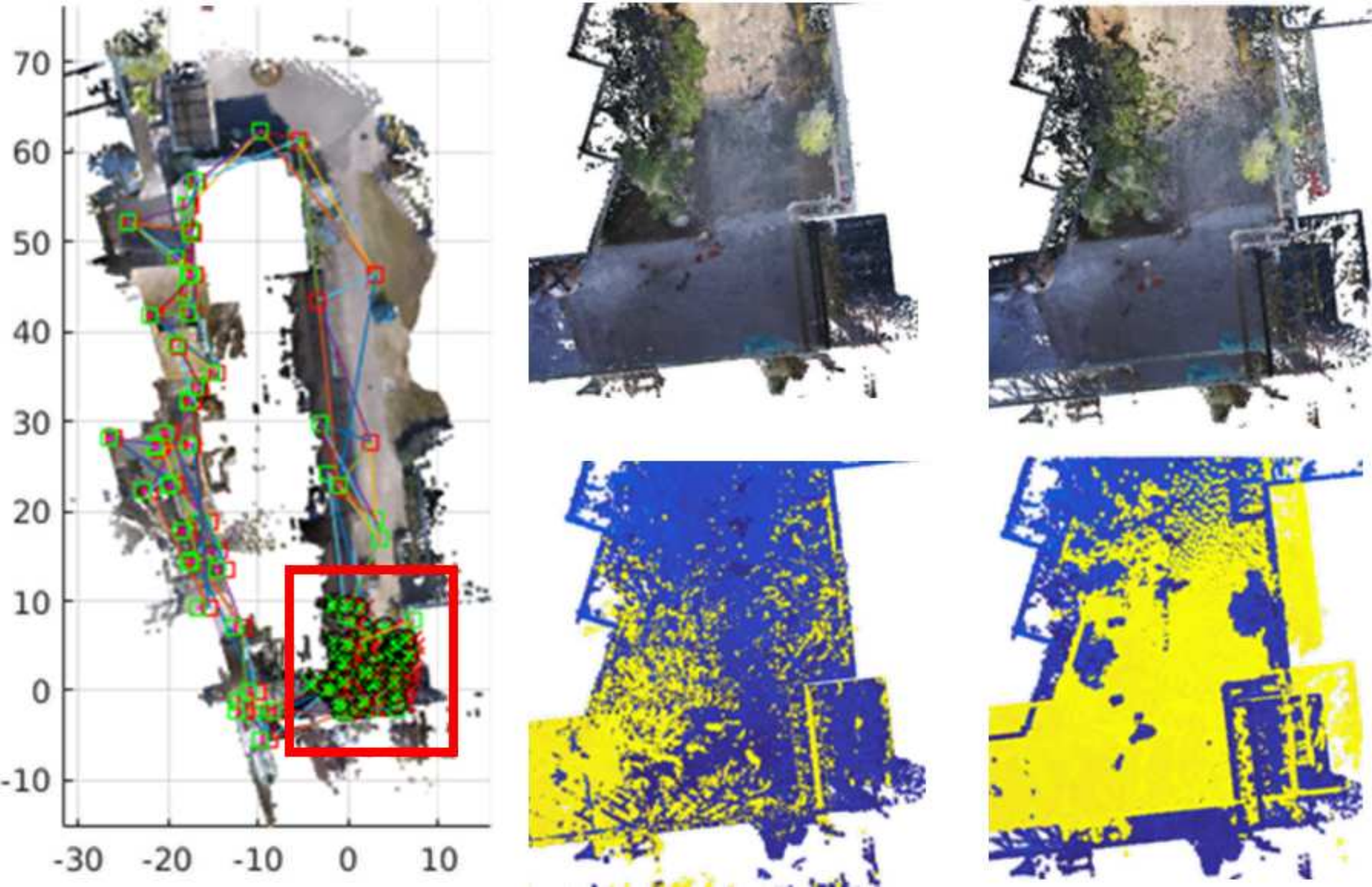}
    \caption{Deformation graph where the estimated misalignment is utilized to close the loop (left). Unit is meter. Before and after comparison of the zoomed loop closure area in red box (middle: after, right: before). Upper figures represent the color map. Lower side color represents time.}
    \label{fig:deformation_result}
\end{minipage} 
\vspace{-2mm}
\end{figure*}

For the evaluation of the localization performance, we have collected \creview{ an indoor and outdoor mixed point cloud dataset that includes multiple loops, a continuous-time trajectory and color images. The collected dataset spans 60$\times$15$m^2$ with
331$m$ in continuous-time trajectory length and 377 seconds recording time.  To demonstrate the algorithm under different challenging
situations, we generated 10 datasets from the map shown in Fig. \ref{fig:deformation_result} by making multiple traverses and picking a different place recognition starting position. The datasets include challenging scenes
where the loop closure sequence starts at geometrically degenerate scenes.}
The ground truth of the alignment is acquired from the globally optimized trajectory. To fairly estimate the robustness against the initial guess, initial poses are randomly generated within a certain amount of uncertainty \creview{(easy, medium and hard in the Table I)} and then a 6 DoF \creview{alignment} is estimated 50 times each at different loop closure location.

\vspace{-2mm}
\subsection{\camrdy{Localization}}

The Root Mean Square Error (RMSE) \creview{between the ground truth and} the estimated alignments is listed in Table \ref{tbl:noise}. 
\camrdy{Sparse surfel-based point-to-plain ICP} is utilized for the sparse ICP (a) which was most vulnerable to incorrect initial guess because of incorrect surfel matchings. The method that combines ICP with visual features (b) showed improved rotation estimation ability but \cblue{initial guesses \camrdy{with a large error} (medium and hard in Table \ref{tbl:noise})} caused occasional failures. 

\creview{For the comparison with the state-of-the-art global ICP registration libraries, PCL (c) and  Open3D (d) are utilized. For both methods, the initial alignments are achieved by 3D feature matchings such as using Signature of Histograms of OrienTations (SHOT)~\cite{tombari2010Daniilidis} (c) or Fast Point Feature Histograms (FPFH)~\cite{rusu2009} (d) and refined with dense point cloud in multi-resolution levels, which makes the alignment estimation less affected by the initial guess. These approaches are similar to method (b) except that multi-modal complementary constraints are jointly optimized in our case.} 

\creview{For both methods (c), (d) the results give reasonably accurate estimations for the places with many geometrical features and enough overlaps. However, in case of (d) the accuracy drops quickly and the estimation diverges especially when the localization occurs in geometrically degenerate scenes or the overlap is not sufficient which causes the failure in estimating a proper initial guess.  On the other hand, the result in (c) shows a constant level of error over different initial guesses which was due to the robustness in the SHOT-based initial pose guess.}  

{Our proposed method \textit{PhotogeoSeq} \creview{(e)} combines the method \creview{Visual+ICP} (b) with the proposed sequential fusion and pose outlier rejection. Its robustness against the initial guess outperforms the state-of-the-art algorithms. However, further improvement was achieved in the translation estimation \creview{as given in the proposed \textit{PhotogeoSeq$^+$} (f) by adding semi-direct features to (e)}. The improvement was due to the increased number of the widely spread visual features. 

Our proposed sequential methods \creview{(e), (f)} utilizes wider area for the localization than the methods (a)-(d). However, rather than utilizing the wide area at a single \creview{alignment} estimation, the proposed method (e), (f) segments the map into frame level and runs independent registrations. This provides a way   of conducting graph-SLAM like loop closure to the map-centric SLAM where the map is continuous and non-rigid rather than discrete like graph-based SLAM.  
}
\camrdy{Fig. \ref{fig:st_map} displays how the continuous map is spatio-temporally segmented into the frame
level for the sequential localization.}

The processing time is given at the last row of Table I in seconds. 
\camrdy{Note that as the localization occurs over different places, sufficient time will be given for processing each localization.} 
Note that the alignment estimation process in the proposed methods (e), (f) will be executed at each $i_{th}$ locations, which is why the processing time is multiplied by the number of matched places $i_{n}$.

\creview{Considering the actual local trajectories are drifting over time, the length of the sequential fusion \camrdy{$i_{n}$} should be limited. Otherwise, the transformed alignments which are found far from the initial loop closure location will be different from the alignment found on the initial location. The maximum length of $i_{n}$ depends on the accuracy of the trajectory estimation and can be found by propagating the unit uncertainty according to the travel time.} 

\vspace{-4mm}
\subsection{\camrdy{Outlier Handling}}

\creview{The pose estimation uncertainty and the error of the fused pose in the sequential fusion is depicted in Fig. \ref{fig:multi_fig} along with the visual evidence test statistics where the color in the $x$ axis label represents the status of the current fusion (black: need more evidence, blue: inlier pose estimation, red: outlier pose estimation). }

The uncertainty and the actual error monotonically decrease whenever a new valid pose estimation is added. Occasional localization failures \camrdy{or false
positive place recognitions}
(numbers in red) are filtered by the localization outlier rejection method.
Once the uncertainty reaches to a threshold,
the result of the localization is utilized for a map deformation as a loop closure constraint \cite{park2017c}. \camrdy{An example of the deformation graph
and the comparison of the registration is depicted in Fig. \ref{fig:deformation_result}.}
\creview{
When the uncertainty of each translation or orientation is high, the pose estimation is not reflected to the fused pose. However, as discussed in Section V.B the measured uncertainty could be incorrect for various reasons. The adverse effect on the rotation estimation such as in the fusion sequence 2, 3 can be caused by incorrect uncertainty.  
}
\creview{Detected inlier features are added to the visual evidence pool and increased
number of evidences are  used for each sequence in the alignment validity test. The inlier rate is calculated by (inlier evidence)$/$(total evidence) where any pose estimations with a large error such as the fusion sequence 6 drastically drop the inlier rate.}

\section{Conclusion}
\label{sec:6}

In this paper, we proposed a robust metric localization method (\textit{PhotogeoSeq$^+$}) that tightly couples spatio-temporal and visual information from a multi-modal sensory set \camrdy{to improve the robustness and accuracy of a loop closure for continuous-time map-centric SLAM. }
Based on our experiments, we demonstrated the proposed method is superior to the state-of-the-art global ICP methods in terms of accuracy and failure detection. This is especially the case when the loop closure is detected for scenes that lack geometric features, as the proposed metric localization method has four times more accurate translation estimation compared to state-of-the-art. 
Furthermore, with the proposed sequential approach where it observes if the estimated alignment is constant over different places, we were able to reject both localization or place recognition failures and achieve robust localization regardless of the initial guess.
Our approach is especially beneficial with the on-the-fly loop closure scheme of the map-centric approach \camrdy{where faultless metric localization
is required.}
Finally, comparison with the state-of-the-art loop closure failure detection method reveals the benefits of our proposed method as it does not require a global trajectory optimization.   
%
%
In our future work, handling local non-rigidity and advanced map deformation that counts multiple alignments will be covered.


\section*{Acknowledgments}
The authors gratefully acknowledge funding of the project by the CSIRO and QUT.
Special thanks goes to Renaud Dube, Pavel Vechersky for the implementation.

\balance

{\small
        \bibliographystyle{IEEEtran}
        \bibliography{ref}
}

\end{document}